\documentclass[sigconf,nonacm]{acmart}

\AtBeginDocument{%
  }

\renewcommand\footnotetextcopyrightpermission[1]{}

\usepackage{amsmath}
\usepackage{booktabs}
\usepackage{multirow}
\usepackage{enumitem}
\usepackage{tabularx}
\newcolumntype{Z}{>{\raggedright\arraybackslash}X}

\newcommand{\loss}{\mathcal{L}}
\newcommand{\ind}{\mathbf{1}}

\begin{document}

\title[TWICE: Two-Clock, Two-Window Learning]{TWICE: Two-Clock, Two-Window Learning for Long-Horizon Conversion Prediction in Online Advertising}

\author{Kaiyuan Li}
\email{likaiyuan03@kuaishou.com}
\affiliation{%
  \institution{Kuaishou Technology}
  \city{Beijing}
  \country{China}}

\author{Kun Wang}
\email{wangkun17@kuaishou.com}
\affiliation{%
  \institution{Kuaishou Technology}
  \city{Beijing}
  \country{China}}

\author{Zhongbo Wang}
\email{wangzhongbo@kuaishou.com}
\affiliation{%
  \institution{Kuaishou Technology}
  \city{Beijing}
  \country{China}}

\author{Teng Sha}
\email{shateng@kuaishou.com}
\affiliation{%
  \institution{Kuaishou Technology}
  \city{Beijing}
  \country{China}}

\author{Ming Yan}
\email{yanming@kuaishou.com}
\affiliation{%
  \institution{Kuaishou Technology}
  \city{Beijing}
  \country{China}}

\author{Yanhua Cheng}
\email{chengyanhua@kuaishou.com}
\affiliation{%
  \institution{Kuaishou Technology}
  \city{Beijing}
  \country{China}}

\author{Xialong Liu}
\email{zhaolei16@kuaishou.com}
\affiliation{%
  \institution{Kuaishou Technology}
  \city{Beijing}
  \country{China}}

\renewcommand{\shortauthors}{Kaiyuan Li et al.}

\begin{abstract}
Long-horizon conversion prediction under delayed feedback creates a two-clock, two-window learning problem in online advertising. A short base observation window releases recent clicks on the click clock before their outcomes mature, whereas conversions continue to arrive on the conversion clock throughout a longer target conversion window. The click clock provides timely but partially observed status supervision. The conversion clock reveals long-tail delays, but the delay composition within an arrival-time slice is weighted by historical click cohorts with different traffic volumes and target-window conversion rates.

We present TWICE, a framework that factorizes long-horizon post-click conversion rate (CVR) into a target-window conversion probability and a grouped elapsed-delay cumulative distribution function (CDF). The two clocks provide complementary supervision. Click-clock records train the target-window CVR head through a current-status likelihood over the base observation window. Newly arrived conversions train the delay model on the conversion clock. To account for the cohort mixture, TWICE uses fixed click-time predicted CVR (pCVR) mass as cohort exposure in an arrival-conditioned likelihood. This accounts for differences in cohort traffic and conversion propensity. The resulting aggregate records are self-contained. A single learned CDF produces monotone predictions for all requested horizons up to the target conversion window. Serving requires neither historical lookup nor convolution. Experiments on a public benchmark and an industrial advertising dataset demonstrate the effectiveness of TWICE. In an online A/B test in Kwai's advertising system, TWICE increased expected revenue, revenue, and conversions by 2.486\%, 1.858\%, and 2.061\%, respectively. It was subsequently deployed to full traffic.
\end{abstract}



\maketitle
\hypersetup{pdfcreator={LaTeX},pdfproducer={pdfTeX}}

\section{Introduction}
Long-horizon conversion prediction under delayed feedback is central to online advertising: users may click now but complete the desired action days or weeks later, so systems must estimate at serving whether it will occur within a long, business-defined target conversion window~\cite{chapelle2014delayed,gu2021defer,dai2023ddfm,su2020postclick,liu2023dfsn,luo2026tesla}.
Post-click conversion rate (CVR) is a core value signal for conversion-aware ranking and post-click outcome estimation~\cite{li2021ftp,lee2012cvr,ma2018esmm,wen2020esm2,wang2022escm2,su2024ddpo}, bid optimization, traffic allocation, and budget-constrained advertising decisions~\cite{zhu2017ocpc,badanidiyuru2022incrementality}.

Production learning is harder: traffic allocation, user intent, active campaigns, and their relationships with conversion outcomes change continuously, requiring daily or more frequent CVR updates.
Each update must learn from recent click cohorts released after a short base observation window, while complete labels arrive only after the longer target conversion window.
Before maturity, unresolved clicks mix true negatives with later positives, and observed status entangles target-window conversion probability with conversion-arrival speed: fresh supervision is incomplete, while complete supervision is stale.

\begin{figure}[t]
\centering
\includegraphics[width=\linewidth]{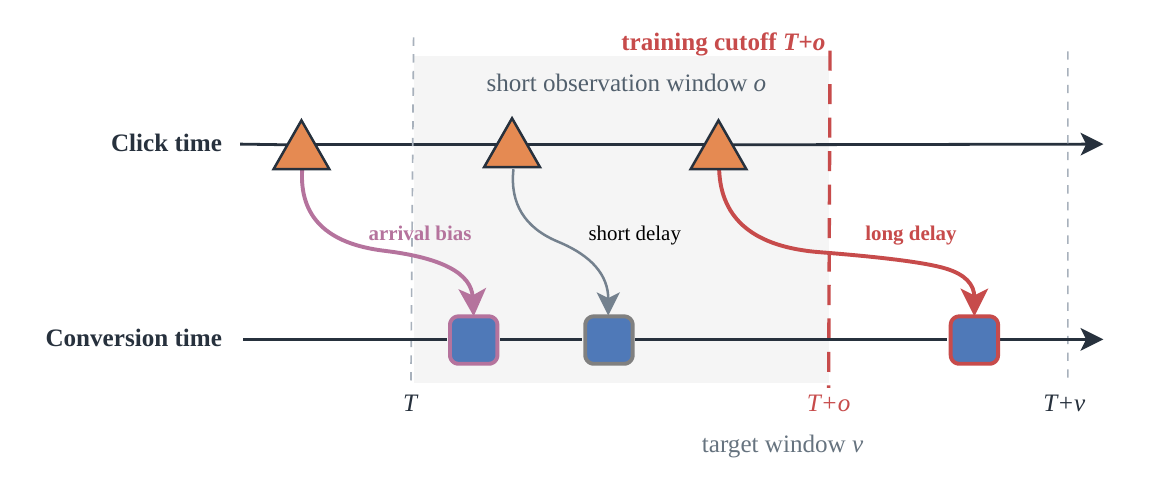}
\caption{Two-clock, two-window view of incremental delayed feedback. In \([T,T+o]\), short-delay conversions are observed by \(T+o\), long-delay positives remain censored, and conversions may arrive from pre-\(T\) clicks, creating cohort-dependent arrival bias.}
\Description{A two-lane timeline shows click time above and conversion time below. The interval from T to the red dashed cutoff T plus o is shaded as the short observation window, while T to T plus v is the target window. Orange triangles mark clicks and blue rounded squares mark conversions. A purple curve connects a click before T to a conversion in the current observation window and is labeled arrival bias. A gray curve shows a recent click whose short-delay conversion is observed before the cutoff. A red curve shows another recent click whose long-delay conversion arrives after the cutoff but before T plus v, leaving that click censored at training time.}
\label{fig:motivation}
\end{figure}

Prior work models censoring and conversion delay~\cite{chapelle2014delayed,yoshikawa2020nodef}, corrects partially observed click labels~\cite{yasui2020feedback,saito2020dual,yasui2022timewindow,wang2023ulc}, uses elapsed-time sampling~\cite{yang2021esdfm}, incorporates delayed examples through additional updates~\cite{gu2021defer,chen2022defuse}, or combines supervision across observation horizons and delay intervals~\cite{li2021ftp,huangfu2022mtdfm,gao2022multihead,liu2024miss}.
Recent methods further distinguish newly observed and delayed sample distributions or approximate their parameter effects~\cite{dai2023ddfm,ding2026ifdfm}, substantially advancing learning from incomplete or subsequently updated click records.

In production incremental training, however, supervision arrives through two event clocks rather than a single click-indexed stream (Figure~\ref{fig:motivation}).
At cutoff \(T+o\), the click clock releases recent records after a bounded observation age, preserving the current traffic distribution and current-cohort features but revealing only partial status by censoring long-delay positives; the conversion clock emits newly arrived conversions throughout the target window, revealing long-tail delays but mixing multiple historical click cohorts.
Consequently, click-clock status combines target-window CVR with observation probability by \(o\), while conversion-clock delay samples are weighted by historical cohort exposure and conversion propensity.

Motivated by this view, TWICE organizes learning around two clocks and windows: \(o\) determines click-clock release, whereas \(v\) defines the target outcome and range of delayed evidence used for learning.
Click-clock records use status at \(o\) and a stopped delay factor to update full-feature CVR for conversion within \(v\); newly arrived conversions within \(v\) update only the delay head.
Because an arrival may come from compatible historical cohorts, its arrival-conditioned likelihood convolves candidate delay probabilities with fixed target-window cohort pCVR mass, improving the factor used on the click clock without replaying historical clicks through the CVR backbone.
To keep the objective non-degenerate and efficient, TWICE conditions delay on a compact click context and CVR on the full feature vector.
Serving evaluates the CVR and small delay heads per click without historical lookup or convolution, yielding monotone CVR estimates for horizons up to \(v\).

On a public delayed-feedback benchmark and a large-scale industrial advertising dataset, we compare target-window CVR with strong baselines and assess conversion-time modeling, traffic dynamics, observation horizons, delay-context granularity, and system cost.
In an online A/B test in Kwai's advertising system serving billions of requests per day, TWICE increased expected revenue, revenue, and conversions by 2.486\%, 1.858\%, and 2.061\%, respectively (all \(p<0.01\)), with unchanged \(36\,\mathrm{ms}\) prediction time; after offline and online validation, it was deployed to full traffic.

Our contributions are threefold.
\begin{itemize}[leftmargin=*,itemsep=1pt,topsep=2pt]
  \item We formulate long-horizon conversion prediction under delayed feedback as a two-clock, two-window learning problem and identify cohort pCVR weighting in incremental conversion-clock supervision.
  \item We introduce an arrival-conditioned temporal-convolution objective combining a grouped delay distribution with fixed historical cohort pCVR mass, so late conversions improve timing estimates without directly updating the CVR backbone.
  \item We develop a grouped incremental implementation with self-contained delay records and no serving-time historical lookup, validated on public and industrial advertising data.
\end{itemize}

\section{Related Work}
Delayed-conversion methods use censored click-indexed examples, late-arriving feedback, or cross-window prediction in online learning.

\noindent\textbf{Click-indexed censored-feedback modeling.}
DFM~\cite{chapelle2014delayed} factors eventual conversion probability and delay, treating recent unconverted clicks as censored rather than negatives.
FSIW~\cite{yasui2020feedback} corrects feedback shift with importance weighting, ULC~\cite{wang2023ulc} applies unbiased delayed-label correction, and ES-DFM~\cite{yang2021esdfm} uses elapsed-time sampling for click-side delayed-label correction.
Other approaches are nonparametric NoDeF~\cite{yoshikawa2020nodef}, dual-learning DLA-DF~\cite{saito2020dual}, and TimeWindow~\cite{yasui2022timewindow}.
TWICE adopts DFM's probability factorization, whereas these methods primarily correct click-indexed censoring or label bias; cohort-weighted delay learning from conversion-clock arrivals has received less attention.

\noindent\textbf{Continuous learning from late-arriving feedback.}
Continuous Training~\cite{ktena2019continuous} and DEFER~\cite{gu2021defer} use additional updates and real-negative replay for late feedback.
DEFUSE~\cite{chen2022defuse} uses unbiased label correction, DDFM~\cite{dai2023ddfm} separates fresh from delayed distributions, and IF-DFM~\cite{ding2026ifdfm} uses influence functions for label reversal.
GDFM~\cite{yang2022gdfm} discounts auxiliary feedback by temporal gap; these methods update or correct click-side CVR with late feedback.
TWICE separates these paths: clicks enter CVR only under the base observation protocol, while later conversions update only the delay head through an arrival-conditioned objective accounting for cohort-exposure bias in their delay composition, an issue that has received comparatively less attention from the conversion-clock perspective.

\noindent\textbf{Window-aware delay prediction.}
FTP~\cite{li2021ftp} models multiple maturity levels, MISS~\cite{liu2024miss} synthesizes predictions across observation intervals, and Personalized Interpolation~\cite{zhang2025personalizedinterpolation} estimates an intermediate window from short- and long-window predictions.
MTDFM~\cite{huangfu2022mtdfm}, Multi-Head~\cite{gao2022multihead}, and bucket-completion methods~\cite{badanidiyuru2021many} predict disjoint or cumulative delayed labels.
These methods motivate monotone outputs but generally require matured, replayed, or teacher-generated supervision; TWICE instead learns the delay kernel from conversion-clock arrivals with an arrival-conditioned likelihood accounting for historical click-cohort conversion intensity, without using those events to supervise CVR.

\section{Problem Formulation}
\subsection{Two Clocks and Two Windows}
For click \(i\), let \(x_i\) be its serving features and \(C_i\) its click time.
Let \(V_i\) be the arrival time of the first matched post-click conversion used by training, and let \(D_i=V_i-C_i\).
Given a long \textbf{target conversion window} \(v\), set \(V_i=D_i=\infty\) if no conversion occurs within \(v\).
The target-window label is
\begin{equation}
Y_i^{(v)}=\ind(D_i\le v),
\end{equation}
which is zero for both non-converting clicks and conversions outside \(v\).
Under a short \textbf{base observation window} \(o<v\), each click is released once on the \textbf{click clock} at \(C_i+o\), with base-window label
\begin{equation}
Y_i^{(o)}=\ind(D_i\le o).
\end{equation}
Matched conversions are observed separately on the \textbf{conversion clock} at \(V_i\), including arrivals from clicks older than the base observation window; they reveal conversion delays but do not create additional CVR-training records for those historical clicks.

\subsection{From Short-Window Observation to Long-Window CVR}
At release age \(o\), fresh click-clock labels identify the base-window conversion probability
\begin{equation}
p_o(x)=P(Y^{(o)}=1\mid x).
\end{equation}
The business target is instead the conversion probability within the long window \(v\):
\begin{equation}
p_v(x)=P(Y^{(v)}=1\mid x).
\end{equation}
To model when a target-window conversion arrives, define its conditional delay cumulative distribution function (CDF)
\begin{equation}
F(\delta\mid x)
=P(D\le\delta\mid Y^{(v)}=1,x),
\qquad 0\le\delta\le v,
\end{equation}
where \(\delta\) is the elapsed time since the click and \(F(v\mid x)=1\).
Because \(o<v\), evaluating this CDF at the release age \(o\) gives
\begin{equation}
\underbrace{p_o(x)}_{\text{short-window CVR}}
=
\underbrace{p_v(x)}_{\text{long-window CVR}}
F(o\mid x).
\label{eq:observation-factorization}
\end{equation}
Here \(F(o\mid x)\) is the fraction of target-window converters observed by age \(o\).
Short-window labels alone cannot distinguish a low long-window conversion probability from slow conversion arrival; conversion-clock supervision is therefore needed to learn this factor.

\subsection{Conversion-Time Sampling Bias}
As Figure~\ref{fig:motivation} illustrates, a conversion arriving in the current observation interval \([T,T+o]\) may be matched to a pre-\(T\) click, so conversion-clock delays mix current and historical click cohorts.
Let \(m(s)\) denote the target-window conversion-mass rate of clicks at calendar time \(s\), obtained by weighting them by \(p_v(x)\), and let \(f(d)\) be the delay density corresponding to \(F\).
Suppressing context conditioning for clarity, the expected conversion-arrival intensity at time \(t\) is the temporal convolution
\begin{equation}
\Lambda(t)=(m*f)(t)=\int_0^v m(t-d)f(d)\,\mathrm{d}d.
\label{eq:arrival-convolution}
\end{equation}
Thus a delay \(d\) observed in the arrival slice at \(t\) is weighted by the compatible cohort mass \(m(t-d)\).
The raw arrival-delay histogram reflects \(f\) only when this mass is stable; changes in traffic or click quality can otherwise change the observed delay mix even if the delay mechanism is unchanged.
Section~\ref{sec:group-convolution} evaluates this convolution at each observed conversion time using grouped historical pCVR mass.

\section{Methodology}
\begin{figure*}[t]
\centering
\includegraphics[width=\textwidth]{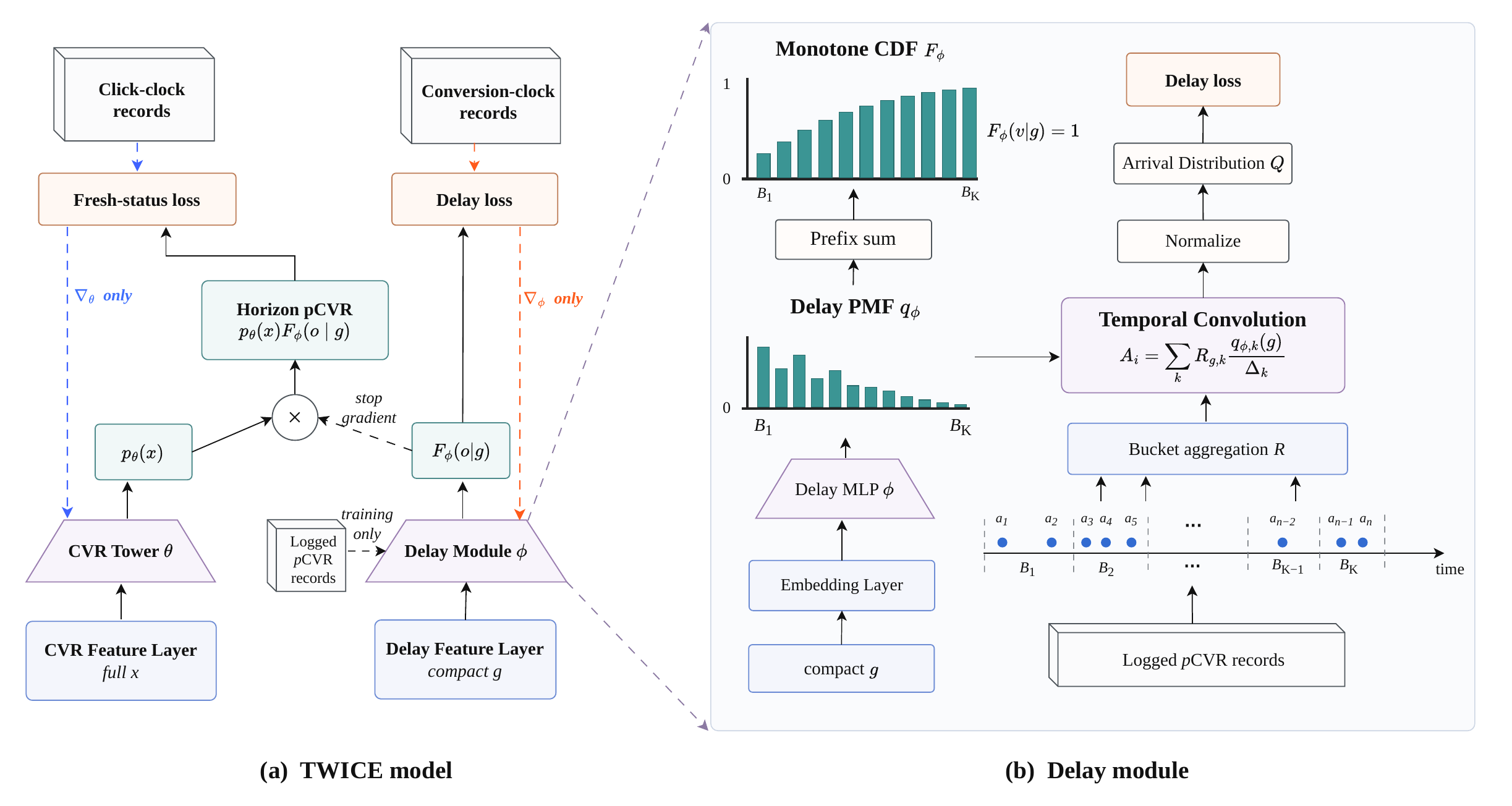}
\caption{Overview of TWICE. (a) Click-clock records train the CVR tower (fresh-status loss) and conversion-clock records train only the delay module; the delay factor in the fresh-status objective is stop-gradient. (b) The delay MLP outputs a PMF over the shared partition \(B_1,\ldots,B_K\); its prefix sums give a monotone CDF, and at each arrival \(V_i\) the same buckets index compatible pCVR masses \(R_i\) that are convolved and normalized into \(Q_i\).}
\Description{A two-panel TWICE diagram. The left panel separates the click-clock CVR branch from the conversion-clock delay branch and shows their gradient paths. The enlarged delay module predicts a PMF over ordered delay buckets B1 through BK, forms a monotone CDF by prefix sum, queries compatible logged pCVR mass at each exact arrival time, and normalizes the temporal convolution into an arrival-conditioned distribution.}
\label{fig:framework}
\end{figure*}

TWICE separates target-window CVR estimation from conversion-time modeling across two clocks.
Click-clock records update the full-feature CVR model once at observation age \(o\), while newly arrived conversions update a compact delay model through arrival-conditioned temporal convolution over historical cohort pCVR mass.
Together, the branches predict CVR at any horizon \(u\le v\) without replaying historical clicks through the CVR backbone.
Figure~\ref{fig:framework} summarizes the architecture and gradient paths; Appendix~\ref{app:notation} summarizes the notation.

\subsection{Click-Clock CVR Learning}
The CVR head predicts whether the click will convert within the target window:
\begin{equation}
p_\theta(x)=\sigma(f_\theta(x)).
\label{eq:cvr-head}
\end{equation}
If all labels were mature, this head could be trained directly with \(Y^{(v)}\).
In online training, however, recent clicks are only partially observed.
TWICE releases each click to the CVR stream once, at the fixed observation age \(o\).
At that time, its fully observed base-window label is \(Y_i^{(o)}\).
TWICE predicts this outcome by
\begin{equation}
\widehat p_o(x_i)
=p_\theta(x_i)\,\mathrm{sg}\!\left[F_\phi(o\mid G(x_i))\right].
\label{eq:fresh-observed-prob}
\end{equation}
For a minibatch's released-click subset \(\mathcal B_{\mathrm c}\), the mean binary cross-entropy (BCE) loss is
\begin{equation}
\begin{aligned}
\loss_{\mathrm{fresh}}(\mathcal B_{\mathrm c})
=-\frac{1}{|\mathcal B_{\mathrm c}|}\sum_{i\in\mathcal B_{\mathrm c}}\!\big[
&Y_i^{(o)}\log\widehat p_o(x_i)\\[-2pt]
&+(1-Y_i^{(o)})\log(1-\widehat p_o(x_i))\big].
\end{aligned}
\label{eq:fresh-loss}
\end{equation}
Without stop-gradient, the product loss can jointly shift \(p_\theta\) and \(F_\phi\), allowing short-window labels to distort the delay CDF.
Stop-gradient instead preserves the two-clock separation: \(F_\phi\) is trained only by timing evidence from the conversion clock, and conversion-clock events do not update the CVR backbone.
Table~\ref{tab:component-ablation} evaluates the latter through \textit{w/o CVR Grad. Isolation}.

\subsection{Grouped Elapsed-Delay Model}
TWICE uses two distinct constructions: \(g=G(x)\) groups clicks with similar delay behavior, while \(B_1,\ldots,B_K\) discretize elapsed time within each group.
For a matched target-window conversion, the observed delay is \(D_i=V_i-C_i\).
Let \(0=b_0<b_1<\cdots<b_K=v\) be an ordered set of delay boundaries.
TWICE partitions \([0,v]\) into
\(B_1=[b_0,b_1]\) and
\(B_k=(b_{k-1},b_k]\) for \(k>1\), with width
\(\Delta_k=b_k-b_{k-1}\).
Non-converters have no delay observation.

Following DFM's feature-conditioned conversion-delay factorization~\cite{chapelle2014delayed}, TWICE retains the full feature vector for target-window CVR but conditions delay on a compact \(G(x)\), built from stable fields such as reporting policy, conversion action, coarse advertiser or industry attributes, channel, device, and coarse click time.
This choice balances timing heterogeneity, well-supported convolutional risk sets, and efficient exact-time range queries; Section~\ref{sec:group-convolution} explains why full-\(x\) conditioning can make the arrival-conditioned objective degenerate.
For delay modeling, we adopt the working approximation
\(F(b_m\mid x)\approx F(b_m\mid G(x))\) for \(m=1,\ldots,K\), whose context granularity is evaluated in Section~\ref{sec:delay-context-selection}.
The delay head predicts the conditional bucket probabilities
\begin{equation}
q_{\phi,k}(g)
=P_\phi(D_i\in B_k\mid g,Y_i^{(v)}=1).
\label{eq:delay-context-sufficiency}
\end{equation}
In implementation \(q_\phi(g)\) is produced by a softmax over the \(K\) elapsed-delay intervals:
\begin{equation}
q_{\phi,k}(g)
=
\frac{\exp(r_{\phi,k}(g))}
{\sum_{\ell=1}^{K}\exp(r_{\phi,\ell}(g))}.
\label{eq:delay-head}
\end{equation}
The delay CDF at each interval edge is the prefix sum of the predicted delay probabilities:
\begin{equation}
F_\phi(b_m\mid g)
=\sum_{k=1}^{m}q_{\phi,k}(g),
\qquad m=1,\ldots,K.
\label{eq:bucket-cdf}
\end{equation}
Inspired by the conversion-CDF interpolation perspective of Personalized Interpolation (PI)~\cite{zhang2025personalizedinterpolation}, TWICE evaluates a requested horizon \(u\in B_m\) with the following piecewise-uniform interpolation:
\begin{equation}
F_\phi(u\mid g)
=\sum_{k=1}^{m-1}q_{\phi,k}(g)
+q_{\phi,m}(g)\frac{u-b_{m-1}}{\Delta_m}.
\label{eq:within-bucket-cdf}
\end{equation}
Combining this CDF with the target-window CVR head gives
\begin{equation}
\widehat p_u(x)
=p_\theta(x)F_\phi(u\mid G(x)),
\qquad u\le v.
\label{eq:serving-score}
\end{equation}
This construction gives horizon-monotone predictions by design:
\[
0\le \widehat p_{b_1}(x)\le\cdots\le
\widehat p_{b_K}(x)=p_\theta(x).
\]
At the bucket level, the same construction gives the joint factorization
\begin{equation}
P_{\mathrm{TWICE}}(Y_i^{(v)}=1,D_i\in B_k\mid x_i)
=p_\theta(x_i)q_{\phi,k}(G(x_i)).
\label{eq:grouped-joint-model}
\end{equation}
This PMF parameterizes the delay kernel but is not fitted directly to the raw arrival-delay distribution.
Section~\ref{sec:group-convolution} instead learns it with an arrival-conditioned temporal-convolution objective.

\subsection{Learning Delay by Temporal Convolution}
\label{sec:group-convolution}
Equation~\ref{eq:arrival-convolution} gives the continuous arrival process.
TWICE evaluates it at every observed conversion's exact arrival time.
For each record \(i\) in a minibatch's conversion-arrival subset \(\mathcal B_{\mathrm a}\), let \(k_i\) be the unique bucket containing the true delay \(D_i=V_i-C_i\).

When historical click \(j\) is scored online, TWICE logs its delay context \(g_j=G(x_j)\), exact click time \(C_j\), and stopped target-window score
\begin{equation}
a_j=\mathrm{sg}\!\left[p_\theta(x_j)\right].
\label{eq:stopped-pcvr}
\end{equation}
Because \(q_\phi\) is conditioned on target-window conversion, \(a_j\) represents expected conversion mass rather than unit click mass.
At the actual arrival time \(V_i\), the compatible pCVR mass for bucket \(k\) is the exact timestamp range sum
\begin{equation}
R_{i,k}
=\sum_{j:g_j=g_i}
a_j\ind(V_i-C_j\in B_k).
\label{eq:routed-pcvr-mass}
\end{equation}
The matched click \(i\) belongs to \(R_{i,k_i}\), so the observed bucket always has positive compatible mass.
Under the piecewise-uniform interpolation in Equation~\ref{eq:within-bucket-cdf}, the corresponding arrival intensity is
\begin{equation}
\Lambda_i
=\sum_{k=1}^{K}
R_{i,k}\frac{q_{\phi,k}(g_i)}{\Delta_k}.
\label{eq:forward-convolution}
\end{equation}
The factor \(1/\Delta_k\) converts bucket probability into a continuous-time density; without it, wider buckets would contribute excess mass.

Conditional on an arrival at \(V_i\), the probability that its delay lies in bucket \(k\) is
\begin{equation}
Q_i(k)
=
\frac{R_{i,k}q_{\phi,k}(g_i)/\Delta_k}
{\Lambda_i}.
\label{eq:group-count-probability}
\end{equation}
TWICE minimizes the arrival-conditioned event NLL
\begin{equation}
\loss_{\mathrm{delay}}(\mathcal B_{\mathrm a})
=-\frac{1}{|\mathcal B_{\mathrm a}|}
\sum_{i\in\mathcal B_{\mathrm a}}\log Q_i(k_i).
\label{eq:delay-loss}
\end{equation}
\paragraph{Assumptions and Scope.}
The objective is exact when logged pCVR mass is calibrated in aggregate within each delay context and click-time range, and within-context converters share a stationary delay distribution over the retained target window.
Production only approximates these conditions: \(G\) is deliberately compact and each click-time score is frozen once logged.
We therefore use Equation~\ref{eq:delay-loss} as a plug-in conditional likelihood that accounts for cohort exposure, rather than claiming universally unbiased delay estimation; Tables~\ref{tab:component-ablation} and~\ref{tab:group-granularity} empirically test sensitivity to both approximations.
Appendices~\ref{app:mass-conservation} and~\ref{app:gradient-equivalence} respectively prove pCVR-mass conservation and bucket/source gradient equivalence for fixed stopped scores, not statistical consistency.
Operationally, each \(R_{i,k}\) is one timestamp range query over stopped scalar pCVR masses; training reads neither historical click features nor the CVR backbone.

\paragraph{Why not a Full-\(x\) Aggregation Key?}
The arrival-conditioned likelihood requires multiple compatible historical clicks in each event-time risk set.
Under exact-key aggregation, setting \(G(x)=x\) makes most high-cardinality risk sets singleton or near-singleton.
For an exact singleton, only \(R_{i,k_i}\) is nonzero, so \(Q_i(k_i)=1\) and the delay loss provides no gradient.
A non-aggregated full-\(x\) likelihood could give every candidate click its own delay distribution, but would require retaining and reevaluating historical click features.
TWICE instead uses a compact \(G(x)\) to preserve cohort competition and self-contained event records; Section~\ref{sec:delay-context-selection} compares the resulting context choices.

\subsection{Joint Training Objective}
\label{sec:joint-training-objective}
TWICE merges the released-click and conversion-arrival streams.
It jointly shuffles their records before sampling mixed minibatches.
For \(\mathcal B=(\mathcal B_{\mathrm c},\mathcal B_{\mathrm a})\), the joint objective is
\begin{equation}
\loss_{\mathrm{TWICE}}(\mathcal B)
=\loss_{\mathrm{fresh}}(\mathcal B_{\mathrm c})
+\lambda_{\mathrm d}\loss_{\mathrm{delay}}(\mathcal B_{\mathrm a}),
\label{eq:twice-loss}
\end{equation}
where each component loss is averaged over its own record subset, and \(\lambda_{\mathrm d}\) controls the relative weight of the delay loss.

With disjoint parameter sets \(\theta\) and \(\phi\), the stop-gradient in Equation~\ref{eq:fresh-observed-prob} and the stopped historical scores in Equation~\ref{eq:stopped-pcvr} give
\[
\nabla_\phi\loss_{\mathrm{fresh}}=0,
\qquad
\nabla_\theta\loss_{\mathrm{delay}}=0.
\]
Consequently, released-click records update the full CVR branch \(\theta\), whereas conversion-arrival records update only the delay branch \(\phi\).
The grouping rule \(G\) is fixed before training.

The stop-gradient in Equation~\ref{eq:fresh-observed-prob} affects optimization only; its forward value equals Equation~\ref{eq:serving-score} evaluated at \(u=o\).
At inference, Equation~\ref{eq:serving-score} produces all supported horizons, and \(F_\phi(v\mid G(x))=1\) gives \(\widehat p_v(x)=p_\theta(x)\).

\subsection{Online Deployment}
\label{sec:online-deployment}
\begin{figure}[t]
\centering
\includegraphics[width=\columnwidth]{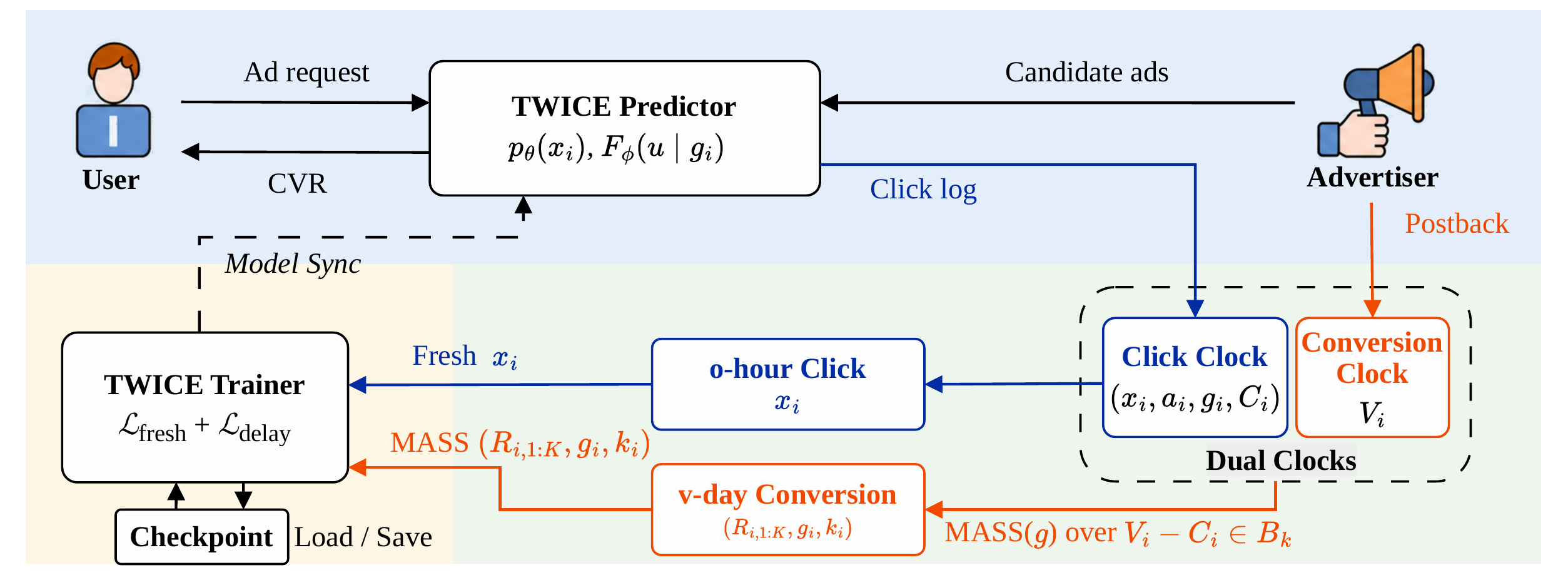}
\caption{TWICE's hourly loop. Clicks log stopped pCVR entries and yield fresh records at age \(o\); conversions materialize self-contained arrival records at \(V_i\). The two streams update their respective branches, and serving reads only the synchronized checkpoint.}
\Description{The online predictor logs click-time pCVR entries to a timestamp-indexed mass state. At each exact conversion arrival time, the materializer queries compatible historical click ranges and creates one self-contained event record. Fresh-click and conversion-arrival records train their respective model branches, and the resulting checkpoint is synchronized back to online serving.}
\label{fig:online-deployment}
\end{figure}

Figure~\ref{fig:online-deployment} shows how TWICE augments the existing CVR system without changing its serving interface.

\textbf{State and records.}
TWICE maintains fresh records \((x_i,Y_i^{(o)})\) released at \(C_i+o\), self-contained arrival records \((g_i,V_i,R_{i,1:K},k_i)\), and a group-indexed timestamp index of stopped pCVR entries \((g_j,C_j,a_j)\).
Only the fresh stream carries full ranking features.
At each exact conversion time \(V_i\), the materializer forms an arrival record from \(K\) compatible range sums over \(V_i-B_k\), without loading historical \(x_j\).
The index retains one scalar entry per unexpired click and evicts it after the event-time watermark passes \(C_j+v\) and all arrivals through that time are materialized.
The grouping rule \(G\), delay partition \(B_{1:K}\), horizon \(v\), and conversion definition are versioned together.

\textbf{Hourly update.}
At bootstrap cutoff \(T_0\), a causal checkpoint uses only observable records; a one-time replay scores clicks in \([T_0-v,T_0]\) to seed the mass index.
Only this bootstrap retains features beyond age \(o\); its scores remain fixed until expiry.
At each later cutoff \(T_n\), training loads the checkpoint from \(T_{n-1}\), consumes only records released in \((T_{n-1},T_n]\), and commits the next checkpoint.
Clicks log stopped scores at \(C_i\) and produce fresh records at \(C_i+o\), while conversions materialize arrival records at \(V_i\).
Cutoffs only batch records: every delay likelihood uses the exact \(V_i\) and \(C_j\), and steady-state scores remain fixed from click time until expiry.

\textbf{Serving.}
Online ranking reads only the synchronized checkpoint: target-window prediction uses \(p_\theta(x)\), and shorter horizons additionally evaluate the lightweight delay head.
Serving never queries the mass index or temporal-convolution state; Table~\ref{tab:online} reports steady-state hourly preparation and training costs and production prediction time.

\begingroup
\setlength{\textfloatsep}{10pt plus 2pt minus 2pt}
\setlength{\floatsep}{8pt plus 2pt minus 2pt}
\setlength{\dbltextfloatsep}{10pt plus 2pt minus 2pt}
\setlength{\dblfloatsep}{8pt plus 2pt minus 2pt}
\section{Experiments}
\label{sec:experiments}
We organize the evaluation around five research questions:
\begin{itemize}[leftmargin=*,itemsep=1pt,topsep=2pt]
  \item \textbf{RQ1:} Does TWICE improve target-window CVR prediction over representative delayed-feedback methods?
  \item \textbf{RQ2:} Which modeling components produce the gain, and how does the granularity of the delay context \(G(x)\) affect predictive performance?
  \item \textbf{RQ3:} How robust is TWICE across base observation windows, requested horizons, and conversion-delay ranges?
  \item \textbf{RQ4:} What qualitative production evidence is consistent with the temporal-convolution mechanism?
  \item \textbf{RQ5:} Do the offline gains translate into production lift with acceptable training and serving cost?
\end{itemize}

\subsection{Experimental Setup}
\textbf{Datasets.}
We use the public Criteo Conversion Logs\footnote{Archived original source: \url{https://web.archive.org/web/20180402081654/http://labs.criteo.com/2013/12/conversion-logs-dataset/}.}~\cite{chapelle2014delayed}, widely used for delayed-feedback CVR prediction~\cite{yasui2020feedback,wang2023ulc,dai2023ddfm}, and one industrial advertising dataset.
We construct chronological click and conversion streams from its click and conversion timestamps.
The industrial dataset contains production clicks matched with delayed conversion events.
The target conversion window is \(v=30\) days for both datasets.
Click traffic spans days 0--60 and conversion logs through day 90, giving every streaming-test click complete 30-day follow-up; post-day-60 events form only retrospective \(Y_i^{(v)}\), never training feedback.
Table~\ref{tab:dataset-statistics} summarizes the dataset scale.

\begin{table}[!t]
\centering
\caption{Statistics of the datasets. Days denotes click-traffic duration, CVR is the 30-day target-window conversion rate, and M denotes million.}
\label{tab:dataset-statistics}
\small
\setlength{\tabcolsep}{4.0pt}
\begin{tabular*}{\linewidth}{@{\extracolsep{\fill}}lccc@{}}
\toprule
Dataset & Days & \#Clicks & CVR \\
\midrule
Criteo & 60 & 15.61M & 22.66\% \\
Industrial Ads & 60 & 174.09M & 0.64\% \\
\bottomrule
\end{tabular*}
\end{table}

\textbf{Implementation.}
Following prior streaming evaluation~\cite{dai2023ddfm}, we split clicks at \(T_0=\) day 30 into a pretraining cohort (days 0--30) and a streaming-evaluation cohort (days 30--60).
Each deployable method is causally initialized from records released by \(T_0\), then makes one chronological pass over its protocol-specific hourly releases before predicting next-hour clicks.
All methods use AdamW, a total minibatch size of \(8192\), and the same dataset-specific backbone: 8-dimensional embeddings with a ReLU MLP \([128,64]\) on Criteo, and the deployed feature encoder and CVR backbone on Industrial Ads.
On Criteo, each method's learning rate is independently selected from \(\{1\!\times\!10^{-6},2.5\!\times\!10^{-6},1\!\times\!10^{-5},5\!\times\!10^{-5},2\!\times\!10^{-4},5\!\times\!10^{-4},1\!\times\!10^{-3}\}\) using a held-out chronological pretraining slice and fixed before streaming evaluation.
Section~\ref{sec:rq-windows} evaluates sensitivity to \(o\); Appendix~\ref{app:twice-configuration} details the delay partition, causal bootstrap, mixed-record sampling, delay-head architecture, and context construction.

\textbf{Baselines.}
Vanilla trains each click once at base observation age \(o\); Oracle follows the same replay cadence using mature \(Y_i^{(v)}\) as its only non-causal information.
We compare the click-indexed methods DFM~\cite{chapelle2014delayed}, FSIW~\cite{yasui2020feedback}, ES-DFM~\cite{yang2021esdfm}, and ULC~\cite{wang2023ulc}; the late-feedback methods DEFUSE~\cite{chen2022defuse}, DDFM~\cite{dai2023ddfm}, and IF-DFM~\cite{ding2026ifdfm}; and the window-aware methods FTP~\cite{li2021ftp}, MISS~\cite{liu2024miss}, and Personalized Interpolation (PI)~\cite{zhang2025personalizedinterpolation}.

\textbf{Metrics and streaming evaluation.}
At cutoff \(T_n\), each replay checkpoint predicts every click in \(\mathcal C_n^{\mathrm{eval}}=\{i:C_i\in(T_n,T_{n+1}]\}\) exactly once, with predictions scored retrospectively against \(Y_i^{(v)}\).
Following prior delayed-feedback evaluations~\cite{yang2021esdfm,dai2023ddfm,ding2026ifdfm}, we report ROC-AUC, average precision as PR-AUC, and mean binary cross-entropy as logloss (LL), computed per next-hour cohort and sample-count-weighted over the streaming period.
For \(M\in\{\mathrm{AUC},\mathrm{PR\mbox{-}AUC},\mathrm{LL}\}\), we also report the relative improvement (RI) between Vanilla and Oracle used in prior evaluation~\cite{dai2023ddfm}:
\begin{equation}
\mathrm{RI}_{M}=
\frac{M-M_{\mathrm{Vanilla}}}
{M_{\mathrm{Oracle}}-M_{\mathrm{Vanilla}}}.
\end{equation}
RI is computed per seed from matched Vanilla and Oracle runs and then averaged; LL reductions give positive RI because both differences are negative.
Across five random seeds, we report means and two-sided \(t\)-tests against the strongest deployable baseline.

\subsection{Overall Performance (RQ1)}
\label{sec:rq-overall}
Table~\ref{tab:overall} compares all deployable methods under a common chronological hourly replay and next-hour evaluation cadence; Oracle only normalizes the recoverable delayed-feedback gap, so we interpret absolute changes first and RI as complementary.

\begin{table*}[t]
\centering
\caption{Overall offline comparison. \(\mathrm{RI}_{\cdot}\) is the recovered percentage of the Vanilla-to-Oracle gain; Oracle is non-deployable and non-causal. Asterisks mark statistically significant improvements over the strongest deployable baseline under a two-sided \(t\)-test (\(p<0.05\)); bold and underlined values are the best and second-best deployable results.}
\label{tab:overall}
\footnotesize
\setlength{\tabcolsep}{1.7pt}
\renewcommand{\arraystretch}{1.12}
\begin{tabular*}{\textwidth}{@{\extracolsep{\fill}}llcccccc@{\hspace{5pt}}cccccc@{}}
\toprule
\multirow{2}{*}{Category} & \multirow{2}{*}{Method} & \multicolumn{6}{c}{Criteo} & \multicolumn{6}{c}{Industrial Ads} \\
\cmidrule(lr){3-8}\cmidrule(lr){9-14}
 & & AUC & \(\mathrm{RI}_{\mathrm{AUC}}\) & PR-AUC & \(\mathrm{RI}_{\mathrm{PR\mbox{-}AUC}}\) & LL & \(\mathrm{RI}_{\mathrm{LL}}\) & AUC & \(\mathrm{RI}_{\mathrm{AUC}}\) & PR-AUC & \(\mathrm{RI}_{\mathrm{PR\mbox{-}AUC}}\) & LL & \(\mathrm{RI}_{\mathrm{LL}}\) \\
\midrule
\multirow{2}{*}{Protocol} & Vanilla & 0.8353 & 0.0\% & 0.6322 & 0.0\% & 0.3996 & 0.0\% & 0.9076 & 0.0\% & 0.1450 & 0.0\% & 0.02545 & 0.0\% \\
 & \textcolor{black!55}{Oracle} & \textcolor{black!55}{0.8426} & \textcolor{black!55}{100.0\%} & \textcolor{black!55}{0.6410} & \textcolor{black!55}{100.0\%} & \textcolor{black!55}{0.3890} & \textcolor{black!55}{100.0\%} & \textcolor{black!55}{0.9226} & \textcolor{black!55}{100.0\%} & \textcolor{black!55}{0.1980} & \textcolor{black!55}{100.0\%} & \textcolor{black!55}{0.02368} & \textcolor{black!55}{100.0\%} \\
\midrule
\multirow{4}{*}{Click-indexed} & DFM~\cite{chapelle2014delayed} & 0.8366 & 17.5\% & 0.6338 & 18.2\% & 0.3970 & 24.2\% & 0.9090 & 9.3\% & 0.1515 & 12.3\% & 0.02536 & 5.1\% \\
 & FSIW~\cite{yasui2020feedback} & 0.8378 & 33.9\% & 0.6350 & 31.9\% & 0.3950 & 43.2\% & 0.9114 & 25.3\% & 0.1570 & 22.6\% & 0.02494 & 28.8\% \\
 & ES-DFM~\cite{yang2021esdfm} & 0.8385 & 43.8\% & 0.6362 & 45.5\% & 0.3939 & 53.8\% & 0.9130 & 36.0\% & 0.1625 & 33.0\% & 0.02476 & 39.0\% \\
 & ULC~\cite{wang2023ulc} & 0.8392 & 53.1\% & 0.6373 & 58.0\% & 0.3928 & 64.0\% & 0.9145 & 46.0\% & 0.1680 & 43.4\% & 0.02457 & 49.7\% \\
\midrule
\multirow{3}{*}{Late-feedback} & DEFUSE~\cite{chen2022defuse} & 0.8388 & 47.6\% & 0.6368 & 52.3\% & 0.3933 & 59.3\% & 0.9105 & 19.3\% & 0.1605 & 29.2\% & 0.02483 & 35.0\% \\
 & DDFM~\cite{dai2023ddfm} & 0.8393 & 54.5\% & 0.6375 & 60.3\% & 0.3925 & 66.9\% & 0.9132 & 37.3\% & 0.1655 & 38.7\% & 0.02466 & 44.6\% \\
 & IF-DFM~\cite{ding2026ifdfm} & 0.8394 & 55.8\% & 0.6376 & 61.4\% & 0.3924 & 67.8\% & 0.9152 & 50.7\% & 0.1750 & 56.6\% & 0.02435 & 62.1\% \\
\midrule
\multirow{3}{*}{Window-aware} & FTP~\cite{li2021ftp} & 0.8379 & 35.3\% & 0.6352 & 34.1\% & 0.3944 & 48.9\% & 0.9108 & 21.3\% & 0.1585 & 25.5\% & 0.02501 & 24.9\% \\
 & MISS~\cite{liu2024miss} & \underline{0.8395} & \underline{57.2\%} & \underline{0.6378} & \underline{63.7\%} & \underline{0.3922} & \underline{69.7\%} & \underline{0.9160} & \underline{56.0\%} & \underline{0.1795} & \underline{65.1\%} & \underline{0.02423} & \underline{68.9\%} \\
 & PI~\cite{zhang2025personalizedinterpolation} & 0.8389 & 49.0\% & 0.6369 & 53.4\% & 0.3932 & 60.2\% & 0.9140 & 42.7\% & 0.1705 & 48.1\% & 0.02448 & 54.8\% \\
\midrule
\textbf{Ours} & \textbf{TWICE} & \textbf{0.8413\raisebox{0.4ex}{*}} & \textbf{81.8\%\raisebox{0.4ex}{*}} & \textbf{0.6388\raisebox{0.4ex}{*}} & \textbf{75.0\%\raisebox{0.4ex}{*}} & \textbf{0.3901\raisebox{0.4ex}{*}} & \textbf{89.7\%\raisebox{0.4ex}{*}} & \textbf{0.9185\raisebox{0.4ex}{*}} & \textbf{72.7\%\raisebox{0.4ex}{*}} & \textbf{0.1880\raisebox{0.4ex}{*}} & \textbf{81.1\%\raisebox{0.4ex}{*}} & \textbf{0.02385\raisebox{0.4ex}{*}} & \textbf{90.4\%\raisebox{0.4ex}{*}} \\
\bottomrule
\end{tabular*}
\end{table*}

TWICE leads all deployable baselines on every metric in both datasets.
Compared with MISS, the strongest baseline, TWICE improves AUC and PR-AUC by \(0.0018\) and \(0.0010\), while reducing LL by \(0.0021\) on Criteo.
On Industrial Ads, the corresponding improvements are \(0.0025\), \(0.0085\), and \(0.00038\); all six are significant at \(p<0.05\).

ULC corrects click-side delayed labels, IF-DFM uses late feedback, and MISS combines observation windows; TWICE additionally models the historical cohort exposure behind each conversion arrival.
Its pCVR-weighted temporal convolution evaluates an arrival against compatible historical cohorts, accounting for traffic-driven cohort mixing in the observed delay composition.
The ablations in Section~\ref{sec:rq-components} verify that both temporal convolution and pCVR weighting contribute to the improvement.
Finally, the absolute gains are substantial relative to the available headroom: on Criteo, the total Vanilla-to-Oracle AUC gap is only \(0.0073\), and TWICE raises \(\mathrm{RI}_{\mathrm{AUC}}\) from MISS's \(57.2\%\) to \(81.8\%\).
On Industrial Ads, its \(0.0085\) PR-AUC gain raises RI from \(65.1\%\) to \(81.1\%\), while \(\mathrm{RI}_{\mathrm{LL}}\) reaches \(90.4\%\).

\subsection{Component Study (RQ2)}
\label{sec:rq-components}
RQ2 isolates the predictive effects of TWICE's modeling choices and delay-context granularity through the following ablations; Appendix~\ref{app:convolution-derivation} derives the exact-time range-sum form.

\subsubsection{Core Modeling Choices and Exposure Robustness}
Under the same chronological replay on both datasets, Table~\ref{tab:component-ablation} compares three one-factor ablations and a fresh-only reference.
\textit{w/o Temporal Conv.} uses direct-arrival cross-entropy; \textit{w/o pCVR Weighting} uses raw click counts instead of cohort pCVR mass; \textit{w/o CVR Grad. Isolation} is an offline diagnostic that lets conversion-clock events update the CVR head and backbone; and \textit{w/o Delay Modeling} is the fresh-only model without a delay head.
For variants with a delay head, Delay-NLL is the mean \(-\log q_{\phi,k_i}(g_i)\) on mature positives; lower is better.

\begin{table}[t]
\centering
\caption{Core ablations and exposure robustness; lower D-NLL is better. Asterisks and bold/underline follow Table~\ref{tab:overall}, with significance tested against the strongest ablation.}
\label{tab:component-ablation}
\scriptsize
\setlength{\tabcolsep}{1.6pt}
\renewcommand{\arraystretch}{1.04}
\resizebox{\linewidth}{!}{%
\begin{tabular}{@{}lcccccc@{}}
\toprule
\multirow{2}{*}{Variant}
& \multicolumn{3}{c}{Criteo}
& \multicolumn{3}{c}{Industrial Ads} \\
\cmidrule(lr){2-4}\cmidrule(lr){5-7}
& AUC & LL & D-NLL & AUC & LL & D-NLL \\
\midrule
\textbf{TWICE} & \textbf{0.8413\raisebox{0.4ex}{*}} & \textbf{0.3901\raisebox{0.4ex}{*}} & \textbf{1.792\raisebox{0.4ex}{*}}
& \textbf{0.9185\raisebox{0.4ex}{*}} & \textbf{0.02385\raisebox{0.4ex}{*}} & \textbf{1.402\raisebox{0.4ex}{*}} \\
\textit{w/o} Temporal Conv. & 0.8397 & 0.3928 & 1.872 & 0.9158 & 0.02430 & 1.566 \\
\textit{w/o} pCVR Weighting & \underline{0.8404} & \underline{0.3917} & 1.824 & \underline{0.9167} & \underline{0.02412} & 1.498 \\
\textit{w/o} CVR Grad. Isolation & 0.8395 & 0.3926 & \underline{1.796} & 0.9153 & 0.02427 & \underline{1.407} \\
\textit{w/o} Delay Modeling & 0.8354 & 0.3995 & -- & 0.9076 & 0.02545 & -- \\
\bottomrule
\end{tabular}%
}
\end{table}

The fresh-only reference nearly matches Vanilla, locating the gains in the delay-aware pathway.
Raw-count convolution is a deliberately severe exposure stress test: removing all pCVR heterogeneity still improves over direct-arrival training by \(0.0007/0.0009\) AUC and \(0.048/0.068\) Delay-NLL on Criteo/Industrial Ads, and exceeds MISS in AUC on both datasets (\(0.8404\) vs.\ \(0.8395\); \(0.9167\) vs.\ \(0.9160\)).
Reintroducing pCVR mass adds \(0.0009/0.0018\) AUC and reduces Delay-NLL by \(0.032/0.096\), showing that TWICE benefits from conversion-propensity weighting without requiring perfectly accurate click-time scores.
Removing CVR-gradient isolation causes the largest AUC drop among delay-aware variants (\(-0.0018/-0.0032\)) but changes Delay-NLL by only \(+0.004/+0.005\), indicating that it protects target-window prediction rather than delay fitting.

\subsubsection{Delay Context Robustness}
\label{sec:delay-context-selection}
Table~\ref{tab:group-granularity} evaluates increasingly refined contexts with a fixed delay partition and \(K\), using field count as a capacity proxy.
Global maps all \(x\) to one key, giving a feasible population-level reference that cannot express context-dependent delay; Minimal \(G_0\) uses six stable ad-side fields, Default \(G\) adds four coarse audience and advertiser fields (10 total), and Expanded \(G_1\) adds ten finer fields from the same families (20 total).
Full-\(x\) is only a no-backoff limit: distinct feature-vector keys fragment compatible risk sets, and an exact singleton has \(Q_i(k_i)=1\) and zero arrival-conditioned gradient.

\begin{table}[t]
\centering
\caption{Sensitivity to delay-context construction on Industrial Ads. Default \(G\) is deployed; bold values denote the best metrics.}
\label{tab:group-granularity}
\footnotesize
\setlength{\tabcolsep}{1.8pt}
\renewcommand{\arraystretch}{1.12}
\begin{tabular*}{\linewidth}{@{\extracolsep{\fill}}lcccc@{}}
\toprule
Delay context & Fields & AUC & Delay-NLL & LL \\
\midrule
Global & 0 & 0.9135 & 1.621 & 0.02470 \\
Minimal \(G_0\) & 6 & 0.9164 & 1.493 & 0.02425 \\
Default \(G\) & 10 & 0.9185 & 1.402 & 0.02385 \\
Expanded \(G_1\) & 20 & \textbf{0.9187} & \textbf{1.387} & \textbf{0.02382} \\
\bottomrule
\end{tabular*}
\end{table}

Metrics improve smoothly from Global to Default \(G\) (\(+0.0050\) AUC, \(-0.219\) Delay-NLL, and \(-0.00085\) LL), whereas doubling the context to Expanded \(G_1\) yields only \(+0.0002\), \(-0.015\), and \(-0.00003\), respectively.
This saturation suggests robustness to moderate context coarsening without a perfectly sufficient grouping function.
We deploy Default \(G\), whose marginal gap to \(G_1\) does not justify twice the context and its smaller, better-supported convolution cohorts.

\begin{table*}[!t]
\centering
\caption{Online A/B test. Business metrics are relative lifts; PCOC and system costs are absolute. Asterisks denote \(p<0.01\).}
\label{tab:online}
\footnotesize
\renewcommand{\arraystretch}{1.00}
\setlength{\tabcolsep}{3.0pt}
\begin{tabular*}{\textwidth}{@{\extracolsep{\fill}}lccccccc@{}}
\toprule
\multirow{2}{*}{Method} & \multicolumn{4}{c}{Business metrics} & \multicolumn{3}{c}{System cost} \\
\cmidrule(lr){2-5}\cmidrule(lr){6-8}
 & \shortstack{Expected\\revenue} & Revenue & Conversions & PCOC & \shortstack{Data\\prep} & \shortstack{Model\\train} & \shortstack{Pred.\\time} \\
\midrule
Production Baseline & -- & -- & -- & 0.961 & 20\,min & 4\,min & 36\,ms \\
\textbf{TWICE} & \textbf{+2.486\%\textsuperscript{*}} & \textbf{+1.858\%\textsuperscript{*}} & \textbf{+2.061\%\textsuperscript{*}} & \textbf{0.990\textsuperscript{*}} & 25\,min & 4\,min & 36\,ms \\
\bottomrule
\end{tabular*}
\end{table*}

\subsection{Window Scaling Study (RQ3)}
\label{sec:rq-windows}
On Criteo, we compare TWICE with the strongest baseline from each family: ULC~\cite{wang2023ulc}, IF-DFM~\cite{ding2026ifdfm}, and MISS~\cite{liu2024miss}.
We vary the base observation age \(o\) at fixed \(v=30\) days and the requested horizon \(u\le v\).
All methods are retrained for each \(o\); for each \(u\), the baselines are retrained, whereas TWICE uses one \(v=30\)-day checkpoint to compute \(\widehat p_u(x)=p_\theta(x)F_\phi(u\mid G(x))\).
Each replay cutoff uses only arrived feedback, with setting-matched Vanilla and Oracle RI anchors.

\begin{figure}[!t]
\centering
\includegraphics[width=\linewidth]{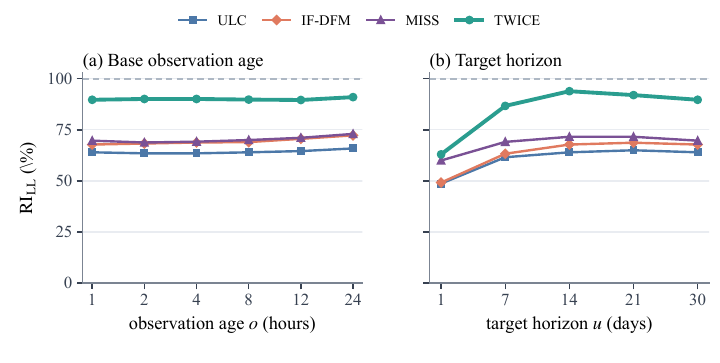}
\caption{Window scaling on Criteo, using setting-matched Vanilla and Oracle anchors. (a) \(\mathrm{RI}_{\mathrm{LL}}\) versus base observation age \(o\); (b) \(\mathrm{RI}_{\mathrm{LL}}\) versus requested horizon \(u\). Baselines are retrained per setting, whereas TWICE uses one \(v=30\)-day checkpoint.}
\Description{Two line charts comparing ULC, IF-DFM, MISS, and TWICE with relative-improvement values as the base observation age and target horizon change.}
\label{fig:window-scaling}
\end{figure}

Across the six observation ages, TWICE recovers \(89.6\%\)--\(91.0\%\) of the Vanilla-to-Oracle LL headroom, varying by only \(1.4\) points.
As \(o\) grows from 1 to 24 hours the baselines improve, since a larger \(o\) exposes more early conversions before label release, but TWICE stays clearly ahead and is far less sensitive to the base cutoff.

For target horizons, the same TWICE checkpoint leads MISS at all five values, and the margin widens beyond one day, indicating that its advantage comes primarily from delayed conversions outside the short window.
Notably, TWICE serves all horizons with a single checkpoint, whereas every baseline must be retrained for each \(u\).

\subsection{Long-Delay Analysis by Delay Time (RQ3)}
Following ULC~\cite{wang2023ulc}, we evaluate five prespecified intervals: \([0,1\mathrm{h}]\), \((1\mathrm{h},1\mathrm{d}]\), \((1\mathrm{d},7\mathrm{d}]\), \((7\mathrm{d},14\mathrm{d}]\), and \((14\mathrm{d},30\mathrm{d}]\).
Let \(P\) be the total number of mature positives and \(P_k\) the number whose realized delay falls in interval \(k\).
Each interval uses the same mature negative pool and weights positives by \(P/P_k\) to preserve full-test CVR; this evaluation-only diagnostic fixes the class prior and isolates converters by delay.
As the strongest deployable baseline in Table~\ref{tab:overall}, MISS represents the baselines alongside Vanilla, TWICE, and Oracle.

\begin{figure}[!t]
\centering
\includegraphics[width=\linewidth]{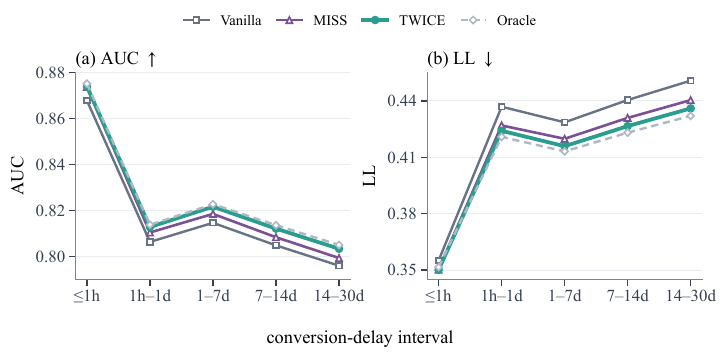}
\caption{Conversion-delay-stratified performance on Criteo. Each interval uses the same mature target-window negative pool, and positive weights preserve the full-test CVR. Higher AUC and lower LL are better.}
\Description{Two line charts compare the AUC and logloss of Vanilla, MISS, TWICE, and Oracle across five fixed conversion-delay intervals.}
\label{fig:long-delay}
\end{figure}

TWICE has the best measured deployable result in both metrics for every interval, and its advantage widens with conversion delay.
Relative to MISS, its AUC gain grows from \(0.00012\) in \([0,1\mathrm{h}]\) to \(0.00405\) in \((14\mathrm{d},30\mathrm{d}]\), while its LL reduction grows from \(0.00004\) to \(0.00441\).
Because the default observation age is one hour, this widening is consistent with TWICE deriving most of its advantage from conversions arriving after the click-clock record has been released.
Oracle is a global non-causal reference rather than a stratum-specific upper bound under this reweighted evaluation, which explains the slight earliest-interval LL crossing.

\subsection{Traffic-Dynamics Case Study (RQ4)}
\label{sec:rq-traffic}
This analysis is intended as a representative production case study.
It uses a traffic-selected launch cohort of advertisers from the same industry.
We align complete advertiser lifecycles at campaign launch and normalize within advertiser, controlling cross-industry timing without selecting a traffic period.
The resulting production-log cohort illustrates the arrival-time bias in Equation~\ref{eq:forward-convolution}.

\begin{figure}[!t]
\centering
\includegraphics[width=\linewidth]{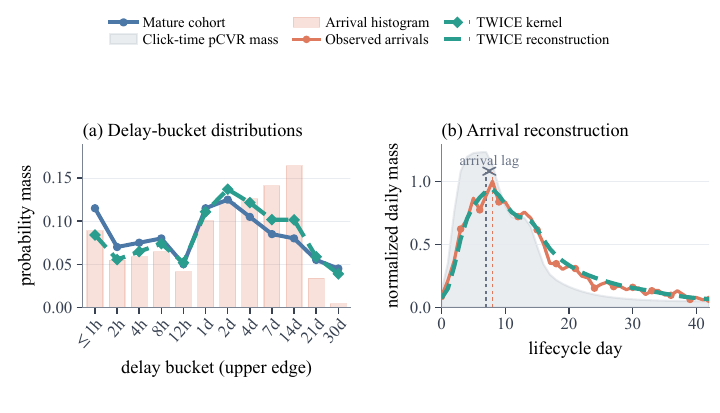}
\caption{Traffic-dynamics case study for an industry launch cohort. Left: arrival histogram, mature click-cohort kernel, and TWICE kernel. Right: observed arrivals and TWICE reconstruction from lifecycle-aligned pCVR mass.}
\Description{Two cohort-aggregated panels comparing arrival and click-cohort delay distributions and showing lifecycle pCVR mass with observed and reconstructed conversion arrivals.}
\label{fig:convolution-case}
\end{figure}

Figure~\ref{fig:convolution-case}(a) visualizes aggregate cohort mixing: the arrival histogram differs from the mature click-cohort distribution, while the TWICE kernel is visually closer to that reference.
In Figure~\ref{fig:convolution-case}(b), convolving the earlier pCVR-mass curve reproduces the main lag and shape of aggregate observed arrivals.
Together, the panels provide qualitative evidence consistent with temporal convolution accounting for traffic-driven mixing.
This mechanistic pattern complements Table~\ref{tab:component-ablation}, where removing convolution or pCVR weighting worsens Delay-NLL and CVR metrics.

\subsection{Online A/B Test (RQ5)}
\label{sec:rq-online}
We compare TWICE with Kwai's incumbent CVR model, an enhanced FSIW-style baseline sharing the same features, conversion definition, deployed backbone, and serving stack.
The eight-day A/B test served \(10\%\) of production traffic; results were computed only after all exposed cohorts matured under the 30-day conversion window.
Expected revenue is the auction's model-based objective, revenue is realized, and PCOC is the ratio of predicted to observed conversions (one indicates aggregate calibration).

TWICE increases expected revenue, revenue, and conversions by 2.486\%, 1.858\%, and 2.061\%, respectively, and improves PCOC from \(0.961\) to \(0.990\); all are significant at \(p<0.01\).
PCOC moves closer to one while all three business metrics improve, showing that the full-system gain is accompanied by better aggregate calibration.
This A/B test measures the full-system effect, while the offline ablations provide component-level attribution.
Prediction remains \(36\,\mathrm{ms}\) and training remains \(4\,\mathrm{min}\) per update; although preparation rises from \(20\) to \(25\,\mathrm{min}\), the \(29\)-minute update pipeline remains within the hourly cadence.
After offline validation and online A/B testing, TWICE was rolled out to full traffic.

\section{Conclusion}
We presented TWICE for long-horizon conversion prediction under delayed feedback as a two-clock, two-window learning problem.
Click-clock records train target-window CVR after a short observation window; conversion-clock events update only the grouped delay model throughout the longer target window.
TWICE learns delay by forward-convolving fixed historical click pCVR mass with the grouped delay distribution at observed conversion times, exploiting long-tail timing and accounting for cohort-weighted arrival bias under the stated calibration and group-stationarity assumptions without contaminating current CVR supervision.
The delay CDF provides monotone predictions up to the target window; serving requires no historical state or convolution, and training retains only an ordered scalar pCVR-mass index, not historical click features.

\endgroup

\clearpage
\bibliographystyle{ACM-Reference-Format}
\bibliography{references}

\appendix
\twocolumn[{%
\begin{minipage}{\textwidth}
\section{Notation Summary}
\label{app:notation}
\centering
\captionof{table}{Summary of recurring notation.}
\label{tab:notation}
\small
\setlength{\tabcolsep}{4pt}
\renewcommand{\arraystretch}{1.14}
\begin{tabularx}{\textwidth}{@{}p{0.16\textwidth}Z p{0.16\textwidth}Z@{}}
\toprule
\textbf{Symbol} & \textbf{Meaning} & \textbf{Symbol} & \textbf{Meaning} \\
\midrule
\(i,j\)
& Focal click or arrival record, and a compatible historical click.
& \(x_i,\;G(x_i)=g_i\)
& Full serving features and the compact delay context of click \(i\). \\
\(C_i,\;V_i,\;D_i\)
& Click time, first matched target-window conversion-arrival time, and delay \(D_i=V_i-C_i\).
& \(o,\;u,\;v\)
& Base observation age, requested prediction horizon, and target conversion window; \(o<v\) and \(u\le v\). \\
\(Y_i^{(w)}\)
& Binary conversion label \(\ind(D_i\le w)\) for \(w\in\{o,v\}\).
& \(p_o(x),\;p_v(x)\)
& Population CVR within the base and target windows. \\
\(p_\theta(x)\)
& CVR-head estimate of the target-window probability \(p_v(x)\).
& \(\widehat p_o(x),\;\widehat p_u(x)\)
& Predicted CVR by release age \(o\) and requested horizon \(u\). \\
\(F(\delta\mid x),\;F_\phi(u\mid g)\)
& Population converter-conditioned delay CDF and its learned grouped approximation.
& \(\theta,\;\phi\)
& Parameters of the CVR and delay branches. \\
\(b_k,\;B_k,\;\Delta_k,\;K\)
& Boundary, interval, width, and number of delay buckets.
& \(q_{\phi,k}(g),\;k_i\)
& Converter-conditioned delay-bucket PMF and the realized bucket of arrival \(i\). \\
\(\mathcal B_{\mathrm c},\;\mathcal B_{\mathrm a}\)
& Released-click and conversion-arrival subsets of a minibatch.
& \(\loss_{\mathrm{fresh}},\;\loss_{\mathrm{delay}}\)
& Released-click BCE and arrival-conditioned delay NLL. \\
\(\loss_{\mathrm{TWICE}},\;\lambda_{\mathrm d}\)
& Joint objective and relative weight of the delay loss.
& \(\mathrm{sg}[\cdot],\;a_j\)
& Stop-gradient and the stopped target-window pCVR mass of click \(j\). \\
\(R_{i,k}\)
& Historical pCVR mass compatible with arrival \(i\) and bucket \(k\).
& \(\Lambda_i,\;Q_i(k)\)
& Model-implied arrival intensity and arrival-conditioned bucket probability. \\
\(m(s),\;f(d),\;\Lambda(t)\)
& Context-suppressed conversion-mass rate, delay density, and arrival intensity.
& \(M_g(I),\;f_\phi(d\mid g)\)
& Group range-summed pCVR mass and learned piecewise delay density. \\
\(\lambda_g(\tau),\;\pi_i\)
& Group arrival intensity and matched-source probability used in the derivations.
& \(T_0,\;T_n\)
& Bootstrap cutoff and the \(n\)-th incremental-update cutoff. \\
\bottomrule
\end{tabularx}
\end{minipage}
}]

\section{Exact-Time Range-Sum Convolution}
\label{app:convolution-derivation}

For historical click \(j\), let \(C_j\) be its click time, \(g_j=G(x_j)\) its delay context, and
\begin{equation}
a_j=\mathrm{sg}\!\left[p_\theta(x_j)\right]
\end{equation}
its stopped target-window pCVR score.
Let \(0=b_0<b_1<\cdots<b_K=v\), \(B_1=[b_0,b_1]\), and \(B_k=(b_{k-1},b_k]\) for \(k>1\).
The interval width is \(\Delta_k=b_k-b_{k-1}\), and the grouped delay PMF satisfies
\begin{equation}
q_{\phi,k}(g)\ge0,
\qquad
\sum_{k=1}^{K}q_{\phi,k}(g)=1.
\label{eq:appendix-pmf-normalization}
\end{equation}
The corresponding piecewise-constant density is
\begin{equation}
f_\phi(d\mid g)
=\sum_{k=1}^{K}
\frac{q_{\phi,k}(g)}{\Delta_k}\ind(d\in B_k).
\label{eq:appendix-delay-density}
\end{equation}

For any click-time interval \(I\), define the additive pCVR mass
\begin{equation}
M_g(I)=\sum_{j:g_j=g}a_j\ind(C_j\in I).
\end{equation}
For a time \(\tau\), let
\(\tau-B_k=\{\tau-d:d\in B_k\}\).
Superposition of the click-level intensities within group \(g\) gives
\begin{align}
\lambda_g(\tau)
&=\sum_{j:g_j=g}a_j f_\phi(\tau-C_j\mid g)
\nonumber\\
&=\sum_{k=1}^{K}
M_g(\tau-B_k)\frac{q_{\phi,k}(g)}{\Delta_k}.
\label{eq:appendix-continuous-convolution}
\end{align}
Evaluating this expression at an observed arrival \(V_i\) gives
\begin{equation}
R_{i,k}=M_{g_i}(V_i-B_k),
\qquad
\Lambda_i=\lambda_{g_i}(V_i),
\end{equation}
which is exactly Equations~\ref{eq:routed-pcvr-mass} and~\ref{eq:forward-convolution}.
Thus the timestamp-indexed pCVR-mass range sum produces the same arrival intensity as summing compatible historical clicks individually.

\section{Conservation of pCVR Mass}
\label{app:mass-conservation}

Within TWICE, \(a_j\) is a fixed target-window pCVR weight, while \(q_{\phi,k}(g_j)\) allocates it across mutually exclusive delay intervals.
Their product marginalizes back to the target-window event:
\begin{equation}
\sum_{k=1}^{K}
P_{\mathrm{TWICE}}(Y_j^{(v)}=1,D_j\in B_k\mid x_j)
=a_j\sum_{k=1}^{K}q_{\phi,k}(g_j)
=a_j.
\label{eq:appendix-joint-marginal}
\end{equation}
The continuous-time intensity preserves the same identity:
\begin{align}
\int_{C_j}^{C_j+v}
a_j f_\phi(\tau-C_j\mid g_j)\,\mathrm d\tau
&=
a_j\sum_{k=1}^{K}
\frac{q_{\phi,k}(g_j)}{\Delta_k}
\int_{B_k}\mathrm d d
\nonumber\\
&=a_j\sum_{k=1}^{K}q_{\phi,k}(g_j)
=a_j.
\label{eq:appendix-mass-conservation}
\end{align}
A historical click may appear in the risk sets of multiple observed arrival times, but those evaluations do not duplicate its probability mass; integrating its density over the target window still yields exactly \(a_j\).
Hourly jobs only batch newly available records, while the likelihood retains the original \(C_j\) and \(V_i\), so update cadence does not alter this identity.

\section{Gradient Equivalence of Source and Bucket Events}
\label{app:gradient-equivalence}

For a conversion-arrival record \(i\in\mathcal B_{\mathrm a}\), the matched click is the source of the arrival at \(V_i\).
Its conditional source probability is
\begin{equation}
\pi_i
=
\frac{a_iq_{\phi,k_i}(g_i)/\Delta_{k_i}}
{\Lambda_i}.
\end{equation}
The bucket probability used by TWICE is
\[
Q_i(k_i)
=
\frac{R_{i,k_i}q_{\phi,k_i}(g_i)/\Delta_{k_i}}
{\Lambda_i}.
\]
Let
\(\ell_{\mathrm{source}}=-\sum_{i\in\mathcal B_{\mathrm a}}\log\pi_i\)
and
\(\ell_{\mathrm{bucket}}=-\sum_{i\in\mathcal B_{\mathrm a}}\log Q_i(k_i)\).
Their difference is
\begin{equation}
\ell_{\mathrm{bucket}}-\ell_{\mathrm{source}}
=\sum_{i\in\mathcal B_{\mathrm a}}\log\frac{a_i}{R_{i,k_i}},
\end{equation}
which contains no \(\phi\) because both \(a_i\) and \(R_{i,k_i}\) are formed from stopped scores.
Hence
\(\nabla_\phi\ell_{\mathrm{bucket}}
=\nabla_\phi\ell_{\mathrm{source}}\).

\section{TWICE-Specific Configuration}
\label{app:twice-configuration}

\textbf{Causal replay and record sampling.}
At \(T_0=\) day 30, the initial checkpoint uses only records observable by \(T_0\).
A click contributes its fresh record once at \(C_i+o\) with the then-observable \(Y_i^{(o)}\), while a conversion-arrival record is released at \(V_i\); thus no \(V_i>T_0\) is exposed during initialization.
Pretraining clicks are excluded from streaming evaluation, but any not-yet-released fresh record is emitted once, their bootstrap pCVR masses remain in the risk index, and later conversions are released at their true arrival times.
At cutoff \(T_n\), TWICE consumes fresh records with \(C_i+o\in(T_{n-1},T_n]\) and arrivals with \(V_i\in(T_{n-1},T_n]\).
The two streams are jointly shuffled at their natural proportions without oversampling, and each hourly update makes one pass using a total minibatch size of \(8192\).

\textbf{Delay model and context.}
TWICE uses \(o=1\) hour and \(\lambda_{\mathrm d}=1\).
For \(v=30\) days, the \(K=12\) delay intervals use boundaries \(0,1,2,4,8,12,24\) hours followed by \(2,4,7,14,21,\) and \(30\) days.
The delay head uses separate embeddings for the fields in \(G\), a two-layer ReLU MLP with widths \([64,32]\), and a \(K\)-way softmax; it shares no trainable parameters with the CVR branch.
On Criteo, Default \(G\) uses all released categorical context columns after excluding timestamps, labels, event identifiers, and continuous cost.
On Industrial Ads, \(G_0\), \(G\), and \(G_1\) contain 6, 10, and 20 stable low-cardinality fields, respectively, from the families described in Section~\ref{sec:delay-context-selection}, with click time coarsened to hour of day.

\end{document}